\documentclass[11pt]{article}

\usepackage[final]{acl}

\usepackage{times}
\usepackage{latexsym}

\usepackage[T1]{fontenc}

\usepackage[utf8]{inputenc}

\usepackage{microtype}

\usepackage{inconsolata}

\usepackage{graphicx}

\usepackage{lipsum}
\usepackage{amsmath}
\usepackage{amssymb}

\usepackage{colortbl} 
\definecolor{LightCyan}{rgb}{0.88,1,1}

\usepackage{booktabs}
\usepackage{multirow}
\usepackage{algorithm}
\usepackage{algpseudocode}
\usepackage{tikz}
\usetikzlibrary{patterns.meta}

\definecolor{figureInk}{HTML}{252525}
\definecolor{figureFrame}{HTML}{555555}
\definecolor{figureGain}{HTML}{E5A86F}
\definecolor{figureDrop}{HTML}{F1D0C7}
\definecolor{figureRecovery}{HTML}{7186A4}
\DeclareRobustCommand{\taskSpecificIcon}{%
  \tikz[baseline=-0.15ex,x=1em,y=1em]{%
    \draw[figureInk,line width=1.5pt]
      (0,0.22)--(1.5,0.22);
    \draw[figureInk,fill=white,line width=0.75pt]
      (0.75,0.22) circle[radius=0.2];
  }%
}

\DeclareRobustCommand{\finetuneIcon}{%
  \tikz[baseline=-0.15ex,x=1em,y=1em]{%
    \draw[figureInk,opacity=.65,line width=1.5pt,
          dash pattern=on 3pt off 2pt]
      (0,0.22)--(1.5,0.25);
    \draw[figureInk,opacity=.65,line width=1.0pt]
      (0.5,0.5)--(1.0,0.0)
      (0.5,0.0)--(1.0,0.5);
  }%
}

\DeclareRobustCommand{\gainIcon}{%
  \tikz[
    baseline=0.15ex,
  ]{%
    \fill[figureGain] (0,0) rectangle (0.50,0.25);
  }%
}

\DeclareRobustCommand{\dropIcon}{%
  \tikz[baseline=0.15ex]{%
    \fill[figureDrop] (0,0) rectangle (0.50,0.25);
    \path[
      pattern={Lines[angle=45,distance=3pt,line width=0.75pt]},
      pattern color=figureFrame
    ] (0,0) rectangle (0.50,0.25);
  }%
}

\DeclareRobustCommand{\recoveryIcon}{%
  \tikz[baseline=0.15ex,x=1em,y=1em]{%
    \draw[figureRecovery,line width=1.5pt]
      (0,0.21)--(1.4,0.21);
    \fill[figureRecovery]
      (0.7,0.6)--(0.3,0.0)--(1.1,0.0)--cycle;
  }%
}

\newcommand{\ie}{{\em i.e.}}
\newcommand{\methodshort}{{CART}}
\newcommand{\methodname}{{Compatible Routing}}
\newcommand{\boldmethodname}{{\textbf{C}omp\textbf{A}tible \textbf{R}ou\textbf{T}ing}}
\newcommand{\moduleZero}{{Paired Projector--LLM Experts}}
\newcommand{\moduleA}{{Distribution-Aware Expert Relevance}}
\newcommand{\moduleB}{{Compatibility-Aware Routing}}
\newcommand{\moduleC}{{Relevance-Guided Expert Consolidation}}

\newcommand{\projexpert}[1]{\mathbf{E}_{\text{p}}^{(#1)}}
\newcommand{\proj}{\text{p}}
\newcommand{\llm}{\text{l}}
\newcommand{\llmexpert}[1]{\mathbf{E}_{\text{l}}^{(#1)}}
\newcommand{\projrelevance}[1]{R_{\text{p}}^{(#1)}}
\newcommand{\llmrelevance}[1]{R_\text{l}^{(#1)}}

\newcommand{\samplex}{\mathbf{x}}
\newcommand{\expertmid}{|\ }

\newcommand{\boldmu}{\boldsymbol{\mu}}
\newcommand{\boldSigma}{\boldsymbol{\Sigma}}

\newcommand{\tinyparagraph}[1]{\noindent\textbf{#1}}

\title{{\methodshort}: {\methodname} for Multimodal Continual Instruction Tuning}

\author{
  Ling-Xiang Zhang,
  Tao Hu,
  Da-Wei Zhou\textsuperscript{\textdagger} \\
  School of Artificial Intelligence, Nanjing University \\
  State Key Laboratory of Novel Software Technology,
  Nanjing University \\
  \textsuperscript{\textdagger} Corresponding author \\
  \texttt{\{zhanglx,hut,zhoudw\}@lamda.nju.edu.cn}
}

\begin{document}
\maketitle
\begin{abstract}

Multimodal Large Language Models (MLLMs) achieve strong performance through instruction tuning, yet real-world applications often require them to continually acquire new capabilities from sequential tasks, \ie, Multimodal Continual Instruction Tuning (MCIT). In typical MLLMs, a projector is employed to align visual representations with the language model. 
However, existing methods primarily protect the language model from forgetting while continuously updating the projector across tasks. Such asymmetric adaptation can cause substantial representation drift in the projector, increasing the mismatch between the projector and language model and consequently exacerbating catastrophic forgetting. 
To address this issue, we propose {{\methodname}} (\methodshort), which explicitly coordinates the adaptation of the projector and language model. Specifically, {\methodshort} maintains task-specific experts for both modules and combines them through weighted mixtures. 
{\methodshort} jointly infers their weights by considering both expert relevance and cross-module compatibility, thereby promoting coordinated adaptation between the two modules. To further control model growth, we introduce relevance-guided expert consolidation, which merges experts exhibiting similar relevance profiles across samples within each module. Extensive experiments demonstrate that {\methodshort} achieves state-of-the-art performance.
\end{abstract}

\section{Introduction}

Multimodal Large Language Models (MLLMs) (\citealp{liu2023visual}, \citealp{dai2023instructblip}, \citealp{wang2025internvl3}, \citealp{bai2025qwen3}, \citealp{qwen35blog}) integrate visual information with the reasoning capabilities of LLMs through cross-modal alignment and multimodal instruction tuning \citep{zhang2026instruction},
establishing strong vision-language understanding.
However, real-world tasks and their data often arrive sequentially rather than all at once~\citep{guo2025comprehensive}. MLLMs must continually acquire new capabilities without forgetting previously learned ones, giving rise to Multimodal Continual Instruction Tuning (MCIT) \citep{chen2024coin},
where catastrophic forgetting caused by cross-task parameter interference remains a central challenge \citep{french1999catastrophic}.

\begin{figure}[t]
  \centering
  \includegraphics[width=\columnwidth]{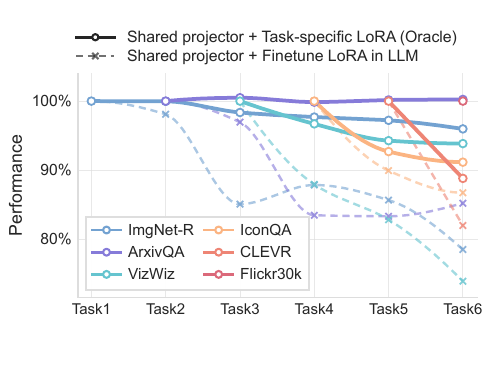}
  \vspace{-11mm}
  \caption{Performance retention during sequential learning. We plot each task's performance, normalized to its initial instruction-tuning peak. Even when the corresponding task-specific LoRA is retrieved, performance on previously learned tasks degrades substantially as the projector is continually updated on new tasks.}
  \label{fig:fig1}
\vspace{-6mm}
\end{figure}

Most MLLMs use a projector to connect a vision encoder to a language model, mapping visual features into the language model's input space~\citep{liu2023visual}.
However, existing MCIT methods mainly mitigate forgetting by protecting language-side adaptations through parameter isolation~\citep{chen2024coin}, selective expansion~\citep{guo2025hide}, or stabilized experts~\citep{xie2026same}, while sharing and continually updating the projector.
Such asymmetric adaptation may alter the visual representations passed to the language model, causing preserved language-side adaptations to process unfamiliar inputs and potentially undermining their ability to retain prior capabilities.

To examine whether language-side protection is sufficient in practice, we study LLaVA-1.5-7B~\citep{liu2024improved} on UCIT~\citep{guo2025hide}, with the vision encoder and pre-trained language-model backbone frozen.
Figure~\ref{fig:fig1} compares two strategies along the same task sequence, both with a shared projector continually updated across tasks: a single LoRA sequentially trained throughout the task stream (\finetuneIcon), and a task-specific LoRA trained for each task as it arrives and retrieved at inference using oracle task identities (\taskSpecificIcon).
Despite retrieving the corresponding task-specific LoRA, performance on earlier tasks still declines as the shared projector evolves, showing that language-side protection alone is insufficient to retain previously acquired multimodal capabilities.

\begin{figure}[t]
  \centering
  \includegraphics[width=\columnwidth]{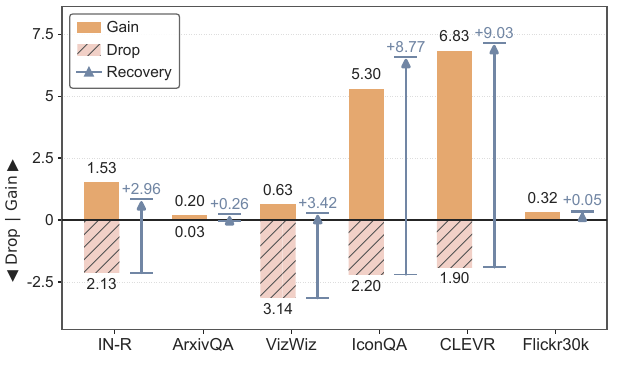}
  \vspace{-5mm}
  \caption{Updating the projector shows a trade-off between learning new tasks and retaining old ones. Although updating the projector improves performance (\gainIcon), learning later tasks can reduce these gains and lower performance on earlier tasks below the frozen-projector baseline (\dropIcon). Retraining the language model with the final projector fixed largely restores performance (\recoveryIcon), suggesting that the mismatch between the projector and language model is indeed an important cause of forgetting.}
  \label{fig:fig2}
\vspace{-4mm}
\end{figure}
Figure~\ref{fig:fig1} shows that updating the shared projector can still cause forgetting even when task-specific language-side adaptations are preserved, making projector freezing a natural alternative.
To investigate whether freezing the projector can mitigate forgetting while adapting to new tasks, 
Figure~\ref{fig:fig2} compares two settings that either continually adapt the shared projector or keep it frozen.
Both settings follow the same task sequence and train a separate language-side LoRA adapter for each task as it arrives.
Each adapter is preserved after training and retrieved at inference using oracle task identities.
In Figure~\ref{fig:fig2}, solid bars (\gainIcon) show performance differences from the frozen-projector baseline immediately after each task is learned, while hatched bars (\dropIcon) show the corresponding differences after the full task sequence, evaluated using the final shared projector and each task's saved LoRA adapter.
As shown in Figure~\ref{fig:fig2}, projector adaptation improves initial performance across the evaluated tasks, but as learning progresses, performance on several previously learned tasks eventually falls below the frozen-projector baseline.
To further examine the source of this degradation, we fix the shared projector at the state reached after completing the entire task sequence and retrain a language-side LoRA adapter for each task using its original training data.
As indicated by the arrows (\recoveryIcon), adapting the language-side parameters to this fixed projector recovers a substantial portion of the lost performance on several tasks.
These results support two conclusions.
{\bf 1)} Adapting the projector improves downstream task acquisition compared with freezing it, but can also introduce forgetting.
{\bf 2)} Mismatch between the updated projector and historical language-side adaptations is a major source of this forgetting.

In this paper, we propose {\boldmethodname} (\textbf{\methodshort}) for multimodal continual instruction tuning. 
Our key idea is to preserve task-specific adaptations for both the projector and language model and jointly determine how these adaptations are reused for each input.
Specifically, {\methodshort} instantiates task-specific experts for the projector and language model, and dynamically composes historical adaptations through weighted expert mixtures. The mixture weights are jointly estimated via coordinate ascent variational inference, considering both distribution-aware expert relevance and projector-language pairing preferences. To prevent unbounded model growth, we further introduce relevance-guided expert consolidation, which merges experts with similar relevance profiles within each module.
Together, these mechanisms enable {\methodshort} to mitigate forgetting while continually adapting to new multimodal tasks.

\section{Related Work}

\tinyparagraph{Multimodal Large Language Models.}
Building on pre-trained visual representations~\citep{radford2021learning}, multimodal large language models (MLLMs) connect visual encoders with language models through cross-modal alignment modules~\citep{liu2023visual,dai2023instructblip}.
Following advances in language-model instruction tuning~\citep{wei2021finetuned,longpre2023flan,zhang2026instruction}, multimodal instruction tuning enables MLLMs to follow diverse user requests~\citep{liu2024improved,dai2023instructblip}.
Their applications cover visual question answering~\citep{goyal2017making}, compositional visual reasoning~\citep{johnson2017clevr,zerroug2022benchmark}, and multimodal generation~\citep{tong2025metamorph}.
Despite this progress, prevailing training pipelines largely rely on predefined datasets and task mixtures~\citep{zhang2024mm,chen2024coin}.
Adapting to new domains, instruction styles, and capabilities through further fine-tuning can interfere with previously learned knowledge~\citep{zhai2023investigating,chen2024coin}, motivating continual adaptation beyond the initial training stage.

\tinyparagraph{Multimodal Continual Instruction Tuning} aims to acquire new capabilities while retaining previously learned knowledge~\citep{chen2024coin}.
Existing approaches mitigate forgetting through regularization, parameter-efficient adaptation, and prompting~\citep{kirkpatrick2017overcoming,wang2023orthogonal,yu2025progressive,zeng2025modalprompt}.
Among expert-based methods, MoELoRA~\citep{chen2024coin} routes inputs among low-rank experts, HiDe-LLaVA~\citep{guo2025hide} combines task-specific expansion with task-general fusion, and SAME~\citep{xie2026same} stabilizes both routing decisions and expert updates.
These studies highlight the importance of preserving and reusing task knowledge under sequential adaptation.
Our {\methodshort} further models distribution-aware relevance and cross-module compatibility to jointly coordinate projector and LLM experts, while bounding model growth through relevance-guided consolidation.
\section{Preliminaries}

\tinyparagraph{Multimodal continual instruction tuning} (MCIT) aims to extend an MLLM's capabilities over sequential tasks while retaining previously acquired knowledge \citep{chen2024coin}.
We consider a task stream $\{\mathcal{D}_t\}_{t=1}^{T}$, where $\mathcal{D}_t=\{(\mathbf{v}_i,\mathbf{q}_i,\mathbf{y}_i)\}_{i=1}^{N_t}$ contains triplets of images, instructions, and answers.
Following the rehearsal-free setting, only $\mathcal{D}_t$ is accessible during stage $t$, and historical training examples are not retained.

A typical MLLM comprises a vision encoder $\phi$, a multimodal projector $\pi$, and a language model.
Given an image $\mathbf{v}$ and instruction $\mathbf{q}$, the input embeddings are $\mathbf{z}=[\pi(\phi(\mathbf{v}));\psi(\mathbf{q})]$, where $\psi$ denotes text embedding and $[\cdot;\cdot]$ denotes sequence concatenation.
The answer $\mathbf{y}=(y_1,\ldots,y_L)$ is generated autoregressively:
\begin{equation}
    p_{\boldsymbol{\Theta}}(\mathbf{y}\mid\mathbf{v},\mathbf{q})
    = \prod_{j=1}^{L}p_{\boldsymbol{\Theta}}(y_j\mid\mathbf{z},\mathbf{y}_{<j}),
    \label{eq:autoregressive}
\end{equation}
where $\boldsymbol{\Theta}$ denotes the tunable parameters, with fixed pre-trained parameters omitted for brevity.
The current-task training loss is:
\begin{equation}
    \mathcal{L}_t(\boldsymbol{\Theta})
    = -\mathbb{E}_{(\mathbf{v},\mathbf{q},\mathbf{y})\sim\mathcal{D}_t}
    \left[\sum_{j=1}^{L}\log p_{\boldsymbol{\Theta}}(y_j\mid\mathbf{z},\mathbf{y}_{<j})\right].
    \label{eq:mcit_loss}
\end{equation}

Under continual learning, the model trains only on current-task data but must retain prior knowledge and perform task-agnostic inference.

\tinyparagraph{Mixture of LoRA Experts}
LoRA~\citep{hu2022lora} represents an update to a frozen weight matrix $\mathbf{W}_0\in\mathbb{R}^{d_{\mathrm{out}}\times d_{\mathrm{in}}}$ as $\Delta\mathbf{W}=s\mathbf{B}\mathbf{A}$, where $\mathbf{A}\in\mathbb{R}^{r\times d_{\mathrm{in}}}$, $\mathbf{B}\in\mathbb{R}^{d_{\mathrm{out}}\times r}$, $r\ll\min(d_{\mathrm{in}},d_{\mathrm{out}})$, and $s=\alpha/r$ is a scaling factor.
MoE-LoRA extends this formulation to a collection of low-rank experts with input-dependent routing.
For a layer input $\mathbf{x}\in\mathbb{R}^{d_{\mathrm{in}}}$, the adapted output is:
\begin{equation}
    \mathbf{h} = \mathbf{W}_0\mathbf{x} + s\sum_{k=1}^{K}\omega_k(\mathbf{x})\mathbf{B}_k\mathbf{A}_k\mathbf{x},
    \label{eq:moe_lora}
\end{equation}
where $K$ is the number of available experts and the routing weights satisfy $\omega_k(\mathbf{x})\geq0$ and $\sum_{k=1}^{K}\omega_k(\mathbf{x})=1$.
A standard learned router computes $\boldsymbol{\omega}(\mathbf{x})=\operatorname{softmax}(\mathbf{W}_g\mathbf{x})$, with $\mathbf{W}_g\in\mathbb{R}^{K\times d_{\mathrm{in}}}$.
This formulation allows conditional combinations of lightweight adaptations while preserving the pre-trained base weights.

\tinyparagraph{Discussions.}
Eq.~\eqref{eq:moe_lora} enables input-dependent reuse of historical adaptations through weighted expert mixtures.
However, an evolving projector $\pi$ can change the visual representations $\pi(\phi(\mathbf{v}))$ received by historical language-side experts.
Preserving these experts alone therefore does not ensure their continued effectiveness.
Effective reuse should account for both expert relevance and cross-module compatibility.
Hence, an ideal model should maintain experts for both the projector and the language model while jointly determining their mixture weights.

\section{Method}
\begin{figure*}[t]
  \centering
  \includegraphics[width=1.0\linewidth]{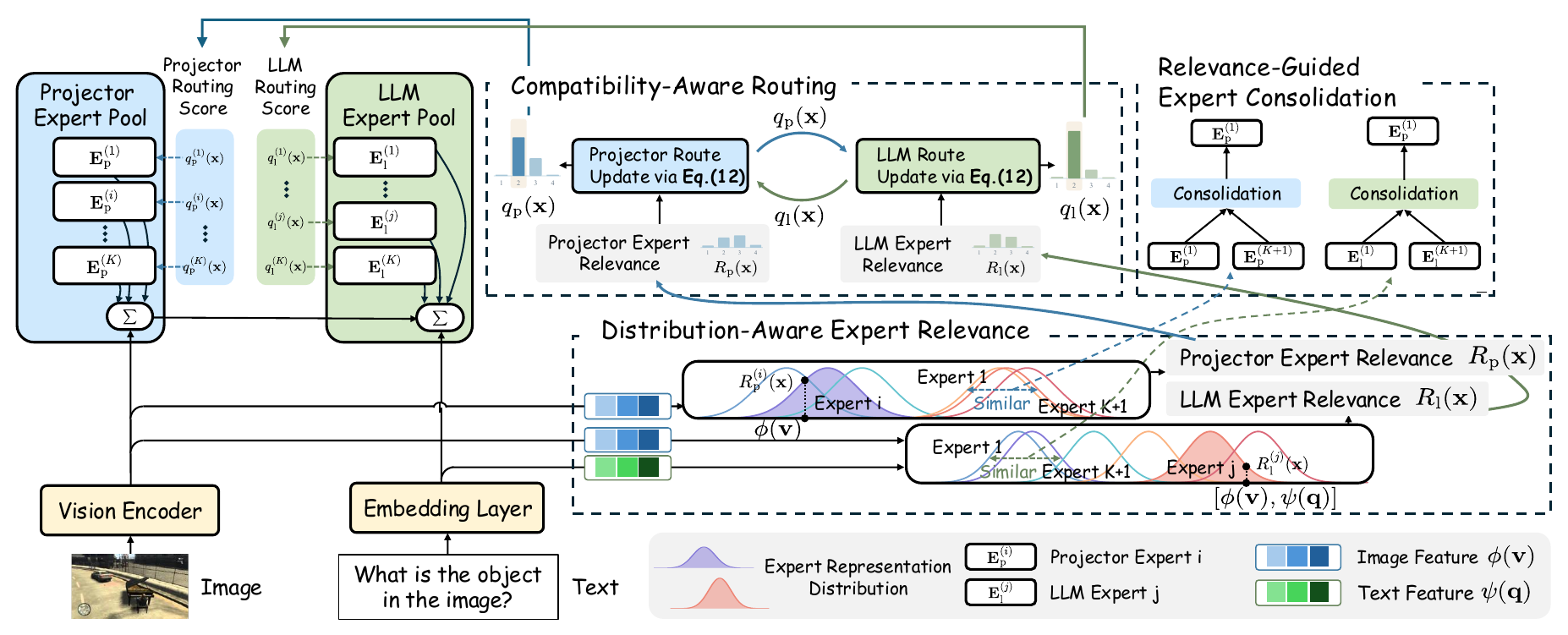}
  \caption{
    Overview of {\methodshort}. We jointly train and preserve expert pairs $(\projexpert{t},\llmexpert{t})$, retain their distribution statistics, and consolidate similar pairs. At inference, Gaussian likelihoods yield relevance scores $R_\text{p}(\samplex)$ and $R_\text{l}(\samplex)$. Alternating compatibility-aware updates produce routing weights $q_\text{p}(\samplex)$ and $q_\text{l}(\samplex)$, which combine projector outputs into a shared visual representation and mix LLM LoRA branches for inference without task identities.
  }
  \label{fig:overview}
  \vspace{-1mm}
\end{figure*}
To mitigate forgetting caused by projector--language model mismatch, we introduce {\methodname} ({\methodshort}) for MCIT.
Figure~\ref{fig:overview} illustrates the overall framework.
Specifically, we maintain projector and LLM expert pools and jointly assign input-specific routing weights to their experts.

\subsection{\moduleZero}
\label{sec:moduleZero}
To prevent forgetting caused by mismatched visual and language adaptations, for each task $t$, we preserve the jointly trained projector and language-side adaptations as a Projector Expert $\projexpert{t}$ with full projector parameters and an LLM Expert $\llmexpert{t}$ parameterized by LoRA. After $T$ tasks, this yields two expert pools $\boldsymbol{\mathcal{E}}_{\proj}=\{\projexpert{1},\cdots,\projexpert{T}\}$ and $\boldsymbol{\mathcal{E}}_{\llm}=\{\llmexpert{1},\cdots,\llmexpert{T}\}$.
At inference time, to enable task-agnostic inference while avoiding cross-module mismatch, we aim to identify relevant and compatible experts for each input. 
For an input $\samplex$, we define the target joint routing distribution $\Pi_{ij}(\samplex) = \Pi(\projexpert{i},\llmexpert{j}\expertmid \samplex)$ as:
\begin{equation}
\Pi_{ij}(\samplex)
=
\frac{
\projrelevance{i}(\samplex)\,
\llmrelevance{j}(\samplex)\,
C_{ij}
}{
\sum_{u, v}
\projrelevance{u}(\samplex)\,
\llmrelevance{v}(\samplex)\,
C_{uv}
},
\end{equation}
where $\projrelevance{i}(\samplex)$ and $\llmrelevance{j}(\samplex)$ denote the relevance of projector expert $\projexpert{i}$ and LLM expert $\llmexpert{j}$ to input $\samplex$, respectively, and $C_{ij}$ characterizes their cross-module compatibility. 

\subsection{\moduleA}
\label{sec:moduleA}

To estimate input to expert relevance without task identities, we characterize each expert by the distribution of its training inputs in a frozen feature space. 
An input is considered more relevant to an expert when its representation better matches the expert's input distribution.

\tinyparagraph{Routing representations.}
For an input $\samplex=(\mathbf{v},\mathbf{q})$, we construct the projector and LLM routing representations as
$\mathbf{h}_{\proj}(\samplex) = \phi(\mathbf{v})$, 
$\mathbf{h}_{\llm}(\samplex) = [\phi(\mathbf{v}), \psi(\mathbf{q})]$. Both feature extractors remain frozen throughout continual learning, ensuring that expert statistics collected at different stages are represented in a consistent feature space.

\tinyparagraph{Expert input distributions.}
To convert the match between an input and an expert into a probabilistic relevance score, we model the expert-conditional routing representations as Gaussian distributions:
\begin{equation}
\begin{aligned}
\mathbf{h}_{\proj}(\samplex)\mid\projexpert{i}
&\sim
\mathcal{N}\left(\boldmu_{\proj}^{(i)},\boldSigma_{\proj}\right),\\
\mathbf{h}_{\llm}(\samplex)\mid\llmexpert{j}
&\sim
\mathcal{N}\left(\boldmu_{\llm}^{(j)},\boldSigma_{\llm}\right),
\end{aligned}
\label{eq:expert_distribution}
\end{equation}
where $\boldmu_{\proj}^{(i)}$ and $\boldmu_{\llm}^{(j)}$ denote expert-specific means that locate the centers of the expert input distributions, while $\boldSigma_{\proj}$ and $\boldSigma_{\llm}$ are shared covariance matrices within the respective expert pools that capture feature variability and correlations.

For each side $m\in\{\proj,\llm\}$, we first compute the feature mean for task $t$ as:
\begin{equation}
\boldmu_m^{(t)}
=
\frac{1}{N_t}
\sum\nolimits_{\samplex\in\mathcal{D}_t}
\mathbf{h}_m(\samplex).
\label{eq:task_mean}
\end{equation}
We retain $\boldmu_m^{(t)}$ as the mean of the corresponding expert, and characterize the within-task feature variation by the scatter matrix:
\begin{equation}
\mathbf{S}_m^{(t)}
=
\sum_{\samplex\in\mathcal{D}_t}
\left(\mathbf{h}_m(\samplex)-\boldmu_m^{(t)}\right)
\left(\mathbf{h}_m(\samplex)-\boldmu_m^{(t)}\right)^{\top}.
\label{eq:task_std}
\end{equation}
The within-task scatter is then aggregated across learned tasks to estimate the shared covariance:
$
\widehat{\boldSigma}_m
=
\sum_{s=1}^{t}\mathbf{S}_m^{(s)}/
\sum_{s=1}^{t}N_s.
$
We then regularize the covariance estimate for numerical stability $\boldSigma_m = \widehat{\boldSigma}_m+\epsilon\mathbf{I}$. As each new task arrives, we update the accumulated scatter and sample count using only its training inputs, without revisiting historical raw data.

\tinyparagraph{Input--expert relevance.}
At inference time, we evaluate the likelihood of the input's routing representation under each expert distribution and normalize these likelihoods using Bayes' rule. For the projector experts, the relevance scores are:
\begin{equation}
\projrelevance{i}(\samplex)
= \frac{1}{Z_{\proj}}
P(\projexpert{i})\,
\mathcal{N}\left(
\mathbf{h}_{\proj}(\samplex);
\boldmu_{\proj}^{(i)},\boldSigma_{\proj}
\right).
\label{eq:relevance_score}
\end{equation}
Here, $Z_\proj$ is the normalization constant given by:
\begin{equation*}
Z_\proj=\sum_{k=1}^{T}
P(\projexpert{k})\,
\mathcal{N}\left(
\mathbf{h}_{\proj}(\samplex);
\boldmu_{\proj}^{(k)},\boldSigma_{\proj}
\right).
\end{equation*}
We compute $\llmrelevance{j}(\samplex)$ analogously using the LLM-side routing representation and distribution statistics. We adopt uniform expert priors because task frequencies at inference time are unknown. 

\tinyparagraph{Discussions.}
By modeling each expert's training-input distribution, we estimate its relevance to the current input without requiring task identities.
The frozen feature space keeps statistics comparable across learning stages, while incremental updates enable rehearsal-free relevance estimation.
These statistics also provide a shared basis for coordinated expert routing and subsequent consolidation.

\subsection{\moduleB}
\label{sec:moduleB}

The target joint routing distribution $\Pi$ accounts for both input--expert relevance and cross-module compatibility. To use this distribution for efficient inference, we require mixture weights for the two expert pools. Specifically, we first combine the projector outputs into a single visual representation, which is then processed by the LLM with weighted LoRA expert branches in a single forward pass. Since all LLM expert branches share this visual representation, the projector mixture weights do not depend on which LLM expert is considered. We therefore approximate the target joint distribution using a factorized routing distribution:
\begin{equation}
q_{ij}(\samplex)
=
q_{\proj}^{(i)}(\samplex)
q_{\llm}^{(j)}(\samplex).
\end{equation}
Although this approximation is factorized, we optimize its two factors jointly to account for cross-module compatibility.

\tinyparagraph{Compatibility-aware approximation.}
Jointly trained projector and LLM experts provide a natural compatibility prior, as their adaptations have been learned together. We encode this preference as:
\begin{equation}
C_{ij}
=
\exp\!\left(\lambda\,\mathbb{I}[i=j]\right),
\end{equation}
where $\lambda \geq 0$ controls the preference for jointly trained expert pairs. This prior favors established pairings while allowing other combinations when supported by input--expert relevance.

We obtain the routing weights by minimizing the KL divergence from the target joint distribution:
\begin{equation}
\label{eq:kl_argmin}
\left(
q_{\proj}^{\star},
q_{\llm}^{\star}
\right)
=
\operatorname*{arg\,min}_{
\substack{
q_{\proj}\in\Delta_{\proj},\ 
q_{\llm}\in\Delta_{\llm}
}
}
D_{\mathrm{KL}}\!\left(
q_{\proj}q_{\llm}
\,\middle\|\,
\Pi
\right),
\end{equation}
where $\Delta_{\proj}$ and $\Delta_{\llm}$ denote the probability simplices over the projector and LLM expert pools, respectively, and we omit the dependence on $\samplex$ throughout this section for readability.

\tinyparagraph{Coordinated routing updates.}
We solve the optimization problem in Eq.~\eqref{eq:kl_argmin} using coordinate ascent variational inference (CAVI)~\citep{blei2017variational}, alternately updating one routing distribution while holding the other fixed.
We initialize the routing weights with the relevance scores:
$
q_{\proj}^{(i)}(\samplex;0)
=
\projrelevance{i}(\samplex),
q_{\llm}^{(j)}(\samplex;0)
=
\llmrelevance{j}(\samplex)$.
At iteration $s$, the coordinate updates are:
\begin{equation}
\begin{aligned}
q_{\proj}^{(i)}(\samplex;s+1)
&\propto
\projrelevance{i}(\samplex)
e^{\lambda q_{\llm}^{(i)}(\samplex;s)},
\\
q_{\llm}^{(j)}(\samplex;s+1)
&\propto
\llmrelevance{j}(\samplex)
e^{\lambda q_{\proj}^{(j)}(\samplex;s+1)}.
\end{aligned}
\label{eq:cavi_update}
\end{equation}
Each update is normalized over the corresponding expert pool. The detailed derivation of these coordinate updates is provided in Appendix~\ref{app:routing_derivation}.

Each update balances an expert's relevance to the input with the current routing weight of its paired expert in the other module.
We perform a fixed number of iterations and use the resulting distributions as the final routing weights.

\tinyparagraph{Expert mixture inference.}
We use $q_{\proj}^{(i)}(\samplex)$ to weight and sum the projector expert outputs, forming a shared visual representation that is concatenated with the instruction embeddings.
The LLM then processes these inputs using the MoE-LoRA formulation in Eq.~\eqref{eq:moe_lora}, with $q_{\llm}^{(j)}(\samplex)$ serving as the expert routing weights at each adapted layer.
This applies the coordinated routing weights within a single LLM forward pass.

\tinyparagraph{Discussions.}
This approximation transforms the compatibility-aware joint distribution over expert pairs into tractable mixture weights for inference.
The coordinated updates allow each module's routing decisions to inform the other, while retaining input--expert relevance as an anchor.
This incorporates cross-module compatibility into expert mixing without separately executing every projector--LLM expert pair.

\subsection{\moduleC}
\label{sec:moduleC}

Introducing a new expert pair for every task causes storage costs to grow with the task sequence.
To keep the expert pool bounded, we impose a capacity of $K_{\max}$ expert pairs.
Whenever this capacity is exceeded, we identify and consolidate the most similar expert pairs using the distribution statistics maintained by Section~\ref{sec:moduleA}.
Projector and LLM experts are consolidated together to maintain the pairing structure used by Section~\ref{sec:moduleB}.

To identify potentially redundant expert pairs, we measure their distributional proximity in both expert pools.
Under the shared-covariance Gaussian model in Eq.~\eqref{eq:expert_distribution}, we compute:
\begin{equation}
d_{\mathrm{m}}(i,j)
=
\left(\boldmu_{\mathrm{m}}^{(i)}
-
\boldmu_{\mathrm{m}}^{(j)}\right)^{\top}
\boldSigma_{\mathrm{m}}^{-1}
\left(\boldmu_{\mathrm{m}}^{(i)}
-
\boldmu_{\mathrm{m}}^{(j)}\right),
\label{eq:distribution_dist}
\end{equation}
where $\mathrm{m}\in\{\mathrm{p},\mathrm{l}\}$ denotes the projector or LLM expert pool.

We select the pair of indices with the smallest weighted combination of projector-side and language-side distribution distances.
Similar input distributions indicate potential expert redundancy and thus guide consolidation.
For the selected indices $u$ and $v$, we define the sample-count weights:
\begin{equation}
    w_u = \frac{N_u}{N_u+N_v},
    \qquad
    w_v = \frac{N_v}{N_u+N_v},
\end{equation}
where $N_u$ and $N_v$ denote the corresponding sample counts.
We merge the projector experts $\projexpert{u}$ and $\projexpert{v}$ into an single expert by averaging their parameters using these sample-count weights:
\begin{equation}
    \boldsymbol{\Theta}_{\proj}^{(\mathrm{new})}
    =
    w_u\boldsymbol{\Theta}_{\proj}^{(u)}
    +
    w_v\boldsymbol{\Theta}_{\proj}^{(v)},
\label{eq:merge_params}
\end{equation}
where $\boldsymbol{\Theta}_{\proj}^{(u)}$ denotes the parameters of $\projexpert{u}$.
For the LLM experts $\llmexpert{u}$ and $\llmexpert{v}$, we combine their weight updates using the same weights $w_u$ and $w_v$, and apply truncated SVD~\citep{eckart1936approximation} to the combined update at each adapted layer to retain the original LoRA rank $r$.

We also update the expert-specific means used for routing:
\begin{equation}
    \boldmu_m^{(\mathrm{new})}
    =
    w_u\boldmu_m^{(u)}
    +
    w_v\boldmu_m^{(v)},
    \quad
    m \in \{\proj,\llm\}.
\label{eq:merge_means}
\end{equation}
The shared covariance $\boldSigma_m$ remains unchanged during
consolidation, as it is estimated from the accumulated within-task
scatter of the original training tasks.
The merged experts $\projexpert{\mathrm{new}}$ and $\llmexpert{\mathrm{new}}$, together with their updated means, replace the original entries under a common index for coordinated routing.
The corresponding sample count is updated to $N_{\mathrm{new}}=N_u+N_v$.
Consolidation is repeated until the number of expert pairs does not exceed $K_{\max}$.

\section{Experiment}

\begin{table*}[t]
\centering
\small
\setlength{\tabcolsep}{2pt}
\renewcommand{\arraystretch}{1.08}
\setlength{\extrarowheight}{1pt}

\label{tab:coin-main}
\resizebox{0.90\linewidth}{!}{
\begin{tabular}{l|cccccccc|c}
\hline
\rowcolor{gray!20}
\textbf{Methods} & ScienceQA & TextVQA & ImageNet & GQA & VizWiz & Grounding & VQAv2 & OCR-VQA & Last \\
\hline

Finetune & 26.00 & 25.38 & 28.51 & 33.07 & 26.52 & 0.10 & 40.00 & 52.92 & 29.06 \\

\rowcolor{gray!10} EWC & 59.75 & 51.06 & 31.88 & 51.84 & 42.26 & 34.96 & 55.30 & 64.55 & 48.95 \\

O-LoRA & 72.56 & 57.66 & 62.84 & 59.14 & 48.43 & 58.97 & 63.21 & 63.31 & 60.77 \\
\rowcolor{gray!10} MoELoRA & 47.34 & 32.91 &38.73 &37.15 & 42.48 & 0.97 &42.77& 57.50&37.48 \\
HiDe-LLaVA & 73.20 & 56.92 & 69.28 & \underline{61.33} & 50.76 & 59.18 & \underline{67.12} & \underline{64.76} & 62.82 \\
\rowcolor{gray!10} ProgLoRA & 74.84 & 51.83 & 83.90 & 49.93 & 53.87 & 31.19 & 62.71 & 64.44 & 59.09 \\
SAME & \underline{78.35} & \underline{60.69} & \underline{90.21} & \textbf{61.70} & \underline{54.13} & \underline{59.87} & 66.04 & 63.59 & \underline{66.82} \\
\hline
\rowcolor{LightCyan} \textbf{\methodshort} & \textbf{84.13} & \textbf{63.85} & \textbf{97.39} & 61.04 & \textbf{61.94} & \textbf{81.08} & \textbf{67.88} & \textbf{66.04} &  \textbf{72.92} \\
\hline
\end{tabular}
}
\vspace{-1mm}
\caption{Main results on the CoIN benchmark with LLaVA-v1.5-7B as the backbone (higher is better). The best and second-best values are marked in \textbf{bold} and \underline{underline}, respectively.}
\label{tab:coin}
\end{table*}
\begin{table*}[t]
\vspace{-2mm}
\centering
\small
\setlength{\tabcolsep}{8pt}
\renewcommand{\arraystretch}{1.08}
\setlength{\extrarowheight}{1pt}

\resizebox{0.90\linewidth}{!}{
\begin{tabular}{l|cccccc|c}
\hline
\rowcolor{gray!20}
\textbf{Methods} & ImgNet-R & ArxivQA & VizWiz & IconQA & CLEVR & Flickr30k & Last \\
\hline
Finetune & 58.03 & 77.63 & 44.39 & 67.40 & 61.77 & \textbf{58.22} & 61.24 \\
\rowcolor{gray!10} EWC & 39.05 & 77.88 & 43.24 & 45.33 & 39.72 & 55.94 & 50.19 \\
O-LoRA & 77.50 & 78.07 & 44.50 & 63.13 & 64.73 & \underline{58.16} & 64.35 \\
\rowcolor{gray!10} MoELoRA & 70.07 & 77.70 & 44.69 & 50.03 & 54.03 & 57.34 & 58.98 \\
HiDe-LLaVA & 84.03 & 90.73 & 44.43 & 58.93 & 41.37 & 54.25 & 62.29 \\
\rowcolor{gray!10} DISCO & 88.88 & \textbf{94.25} & 47.52 & 69.50 & 60.75 & 56.32 & 69.54 \\
SAME & \underline{89.91} & \underline{91.40} & \textbf{55.33} & \underline{77.51} & \underline{68.85} & 55.43 & \underline{73.07} \\
\hline
\rowcolor{LightCyan} \textbf{\methodshort} & \textbf{90.93} & 91.07 & \underline{47.73} & \textbf{88.00} & \textbf{81.20} & 55.12 & \textbf{75.67} \\
\hline

\end{tabular}
}
\vspace{-1mm}
\caption{Main results on the UCIT benchmark with LLaVA-v1.5-7B as the backbone (higher is better). The best and second-best values are marked in \textbf{bold} and \underline{underline}, respectively.}
\label{tab:ucit}
\vspace{-2mm}
\end{table*}

\subsection{Experiment Setup}

\tinyparagraph{Datasets.}
We evaluate on CoIN~\citep{chen2024coin} and UCIT~\citep{guo2025hide}, two benchmarks for multimodal continual instruction tuning.
CoIN covers eight tasks, namely ScienceQA~\citep{lu2022learn}, TextVQA~\citep{singh2019towards}, ImageNet~\citep{deng2009imagenet}, GQA~\citep{hudson2019gqa}, VizWiz~\citep{gurari2018vizwiz}, Grounding~\citep{kazemzadeh2014referitgame,mao2016generation}, VQAv2~\citep{goyal2017making}, and OCR-VQA~\citep{mishra2019ocr}.
These tasks span question answering, image classification, and referring expression grounding.
UCIT comprises ArxivQA~\citep{li2024multimodal}, CLEVR-Math~\citep{lindstrom2022clevr}, IconQA~\citep{lu2021iconqa}, ImageNet-R~\citep{hendrycks2021many}, VizWiz-caption~\citep{gurari2020captioning}, and Flickr30k~\citep{plummer2015flickr30k}.
It supplements CoIN with tasks selected to reduce overlap with LLaVA's original training data, including image captioning.
Detailed dataset statistics for both benchmarks are provided in Appendix~\ref{app:dataset_details}.

\tinyparagraph{Comparison Methods.}
We compare sequential fine-tuning, 
EWC~\citep{kirkpatrick2017overcoming}, 
O-LoRA~\citep{wang2023orthogonal}, 
MoELoRA~\citep{chen2024coin}, 
HiDe-LLaVA~\citep{guo2025hide}, ProgLoRA~\citep{yu2025progressive}, SAME~\citep{xie2026same} and DISCO~\citep{guo2025federated}.

\tinyparagraph{Training Setup.}
Following Prism~\citep{tang2026prism}, we use LLaVA-v1.5-7B~\citep{liu2024improved} on CoIN~\citep{chen2024coin} and UCIT~\citep{guo2025hide}, with a CLIP ViT-L/14 encoder ($336\times336$ inputs)~\citep{radford2021learning} and a two-layer GELU projector.
We train projector experts and LLM LoRA experts ($r=16$, $\alpha=32$), keeping the vision encoder and base LLM frozen.
Each task uses one epoch of AdamW with projector/LoRA learning rates of $5\times10^{-5}/2\times10^{-4}$ and a $0.03$ warmup ratio.
Training is conducted on four RTX 4090 GPUs with a per-device batch size of $4$, $2$ gradient accumulation steps, DeepSpeed ZeRO-2, and \texttt{bf16} precision.
Unless otherwise specified, we use $\lambda=2$, $S=10$ CAVI iterations, capacity $K_{\max}=5$ expert pairs, and covariance regularization $\epsilon=0.1$.
We use PyTorch 2.8.0~\citep{paszke2019pytorch} and pycocotools 2.0.11~\citep{lin2014microsoft}.

\tinyparagraph{Evaluation Metrics.}
Following CoIN~\citep{chen2024coin} and Prism~\citep{tang2026prism}, we report \textit{Last Accuracy} (Last) and \textit{Average Accuracy} (Avg):
$\mathcal{A}_{\mathrm{Last}}=\frac{1}{T}\sum_{s=1}^{T}\mathcal{A}_{s,T}$ and
$\mathcal{A}_{\mathrm{Avg}}=\frac{1}{T}\sum_{t=1}^{T}\frac{1}{t}\sum_{s=1}^{t}\mathcal{A}_{s,t}$,
where $\mathcal{A}_{s,t}$ is the accuracy on task $s$ after learning $t$ tasks, and $T$ is the total number of tasks.
Last evaluates final-stage performance, while Avg averages seen-task performance across stages.
\begin{table*}[t]
\centering
\small
\setlength{\tabcolsep}{7pt}
\renewcommand{\arraystretch}{1.10}
\resizebox{0.96\linewidth}{!}{
\begin{tabular}{l|cccccc|cc}
\hline
\rowcolor{gray!20}
\textbf{Variants}
& ImgNet-R & ArxivQA & VizWiz
& IconQA & CLEVR & Flickr30k
& Last & $\Delta$ \\
\hline

\multicolumn{9}{l}{
\textit{Component Ablation}
} \\
\hline

w/o Distribution-Aware Relevance
& 80.27 & 91.60 & 46.38
& 79.07 & 78.67 & 54.06
& 71.68 & -3.99 \\

\hline
\multicolumn{9}{l}{
\textit{Routing Strategy Comparison}
} \\
\hline
Hard Routing
& 73.87 & 91.73 & 45.88 & 71.73 & 73.73 & 52.80 & 68.29 & -7.38 \\

Independent Routing
& 85.07 & \textbf{92.40} & 46.41 & 71.20 & 69.60 & 54.53 & 69.87 & -5.80 \\

Naive Joint Routing
& 86.33 & 91.33 & 45.99 & 76.33 & 75.67 & \textbf{55.13} & 71.80 & -3.87 \\

\rowcolor{LightCyan}
\textbf{{\methodshort} (Full)}
& \textbf{90.93} & 91.07 & \textbf{47.73}
& \textbf{88.00} & \textbf{81.20}
& 55.12 & \textbf{75.67} & -- \\

\hline
\end{tabular}
}
\vspace{-1mm}
\caption{
Ablation study on the UCIT benchmark.
$\Delta$ denotes the average accuracy change
relative to full {\methodshort}. Best results are
highlighted in \textbf{bold}.
}
\label{tab:ablation}
\end{table*}

\begin{figure*}[!t]
  \centering
  \vspace{-2mm}
  \includegraphics[width=1.00\linewidth]{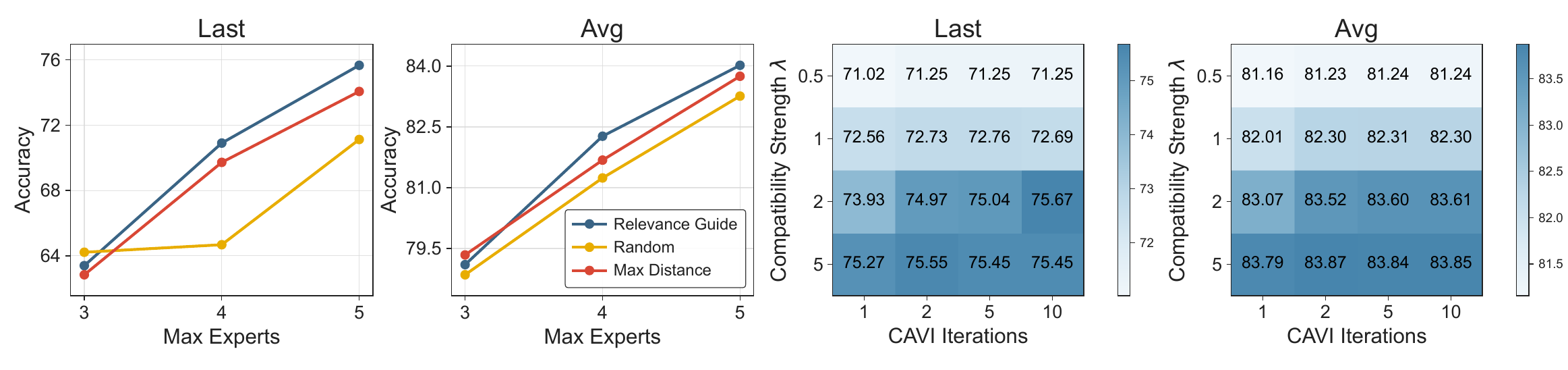}
  \vspace{-6mm}
  \caption{Analysis of expert-pool capacity and CAVI hyperparameters on the six-task UCIT benchmark. The left panels compare consolidation strategies under different expert-pool limits, while the right panels examine compatibility strength and the number of CAVI iterations. Overall, stronger compatibility coupling improves retention, and performance stabilizes after only a few CAVI iterations.}
  \label{fig:max_experts}
\end{figure*}

\begin{figure*}[!t]
\vspace{-3mm}
  \centering
  \includegraphics[width=1.0\linewidth]{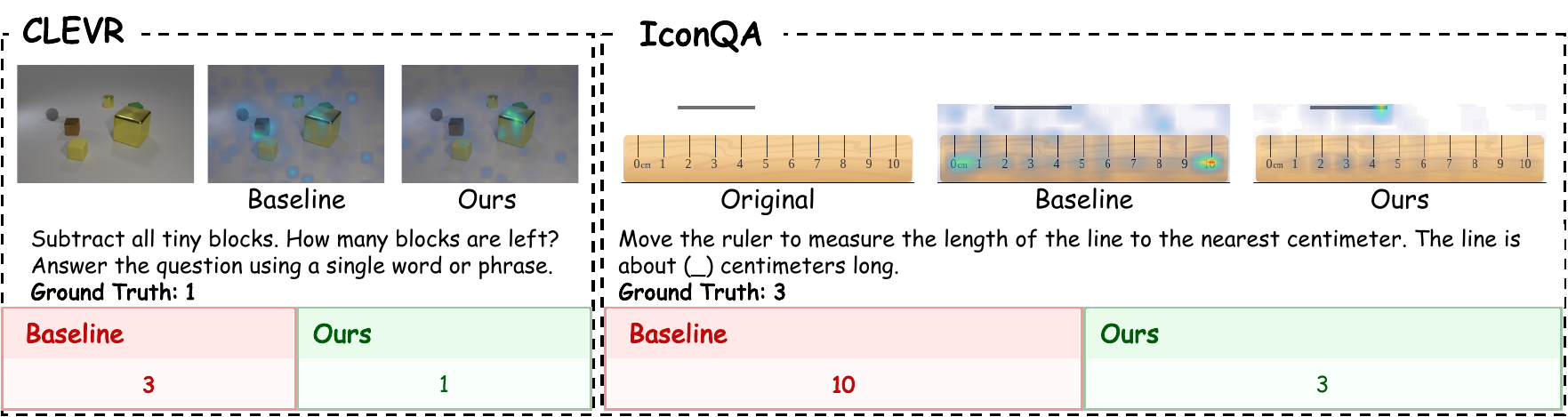}
  \caption{Qualitative comparison with NCM-based routing. {\methodshort} better attends to instruction-relevant visual evidence and produces correct answers.}
  \label{fig:case}
  \vspace{-2mm}
\end{figure*}

\subsection{Main Results}

Tables~\ref{tab:coin} and~\ref{tab:ucit} summarize the main results on CoIN and UCIT. On CoIN, {\methodshort} obtains the highest Last Accuracy of 72.92 and exceeds the strongest baseline SAME by 6.10 points. {\methodshort} also ranks first on UCIT, reaching 75.67 and improving over SAME by 2.60 points. Its leading performance across both the standard CoIN benchmark and the cleaner UCIT setting shows that the gains remain consistent under different task compositions and data characteristics.

\subsection{Ablation Study}

Table~\ref{tab:ablation} evaluates distribution-aware relevance and routing strategies on UCIT.
Removing distribution-aware relevance lowers average accuracy from 75.67 to 71.68, with the largest drops on ImageNet-R and IconQA, supporting input--expert distribution matching.
To assess the effectiveness of compatibility-aware routing, we compare three routing alternatives:
Hard Routing independently selects each module's most relevant expert;
Independent Routing directly uses the relevance distributions as mixture weights, i.e., $q_m(\samplex)=R_m(\samplex)$ for $m\in\{\proj,\llm\}$, without cross-module coordination.;
Naive Joint Routing assigns identical routing weights to both modules by normalizing the element-wise product of their relevance distributions, i.e., $q_{\proj}(\samplex)=q_{\llm}(\samplex)\propto R_{\proj}(\samplex)\odot R_{\llm}(\samplex)$.
Their average accuracies are 68.29, 69.87, and 71.80, respectively, versus 75.67 for \methodshort, demonstrating the benefit of cross-module compatibility in coordinating projector and LLM routing.
\subsection{Further Analysis}

\tinyparagraph{Effectiveness of {\moduleC}.}
On six-task UCIT, Figure~\ref{fig:max_experts} compares consolidation strategies under different expert-pool limits $K_{\max}$. At $K_{\max}=4$, our strategy achieves 70.9 final-stage and 82.3 average accuracy, outperforming random consolidation by 6.2 and 1.0 points, respectively. This advantage persists at $K_{\max}=5$, while all strategies perform similarly at $K_{\max}=3$ due to limited consolidation flexibility. These results show that merging experts with similar projector- and language-side distributions better preserves task knowledge within a bounded expert pool.

\tinyparagraph{Qualitative Visualization.}
Figure~\ref{fig:case} compares {\methodshort} with an NCM-based baseline that retains the LoRA-MoE architecture but replaces {\moduleA} and {\moduleB} with nearest-class-mean routing. 
On CLEVR, {\methodshort} identifies the blocks relevant to subtraction and answers one instead of three. 
On IconQA, it attends near the correct ruler endpoint and predicts three centimeters rather than ten. These examples show that relevance estimation and expert coordination improve routing and preserve instruction-relevant cross-modal alignment.

\section{Conclusion}
This work investigates projector--language mismatch as a source of forgetting in multimodal continual instruction tuning.
We introduce \methodshort, which preserves paired projector and LLM experts and jointly routes them using input--expert relevance and cross-module compatibility for task-agnostic inference.
Relevance-guided consolidation bounds model growth by merging redundant expert pairs.
Experiments on CoIN and UCIT demonstrate superior performance, with ablations supporting the effectiveness of relevance estimation and coordinated routing.
These results highlight the importance of cross-module compatibility for continual multimodal adaptation.

\label{sec:bibtex}

\section*{Limitations}
While {\methodshort} demonstrates strong performance on CoIN and UCIT, it has not yet been evaluated on substantially longer task sequences.
Extending evaluation to longer continual streams would provide a more comprehensive assessment of long-term knowledge retention and remains an important direction for future work.

\section*{Ethical Considerations}

{\methodshort} may inherit biases, harmful behaviors, and
hallucinations from pretrained models and training data.
Distribution-based routing may be less reliable for
underrepresented inputs, potentially affecting users of
applications such as visual assistance. Downstream deployment
therefore requires fairness and safety evaluation after model
updates, along with appropriate human oversight.

More adaptable multimodal models could be misused for
privacy-invasive image analysis or misleading content generation.
Although our routing mechanism does not revisit historical
raw data, retaining aggregate statistics and learned experts
does not provide a formal privacy guarantee. Sensitive
applications require appropriate data governance and
privacy assessment.

Training and evaluation consume energy. Low-rank adaptation
and expert consolidation control trainable parameters and
expert-pool growth, but do not necessarily reduce overall
energy consumption or carbon emissions.

\bibliography{custom}

\appendix

\section{Dataset Details}
\label{app:dataset_details}

Tables~\ref{tab:coin_dataset_details} and~\ref{tab:ucit_dataset_details} summarize the domains and split sizes of the datasets in CoIN and UCIT, respectively.
Train and Test denote the numbers of samples in the corresponding splits.

\tinyparagraph{CoIN.}
CoIN contains eight datasets spanning image classification, visual question answering, referring expression comprehension, visual assistance, and scientific reasoning.
Its training sets contain between 20,000 and 34,602 samples, while each test set contains 3,000 samples.

\begin{table}[H]
\centering
\small
\begin{tabular*}{\columnwidth}{@{\hspace{\tabcolsep}\extracolsep{\fill}}llrr@{\hspace{\tabcolsep}}}
\toprule
\textbf{Dataset} & \textbf{Domain} & \textbf{Train} & \textbf{Test} \\
\midrule
GQA & Compositional VQA & 20,000 & 3,000 \\
ImageNet & Image classification & 20,000 & 3,000 \\
TextVQA & Scene-text VQA & 34,602 & 3,000 \\
OCRVQA & OCR-based VQA & 20,000 & 3,000 \\
REC-COCO & Referring expressions & 20,000 & 3,000 \\
VizWiz & Visual assistance & 20,523 & 3,000 \\
VQAv2 & General VQA & 20,000 & 3,000 \\
ScienceQA & Science reasoning & 20,000 & 3,000 \\
\bottomrule
\end{tabular*}
\caption{Dataset statistics for the eight-task CoIN benchmark.}
\label{tab:coin_dataset_details}
\end{table}

\tinyparagraph{UCIT.}
UCIT contains six datasets spanning artistic image recognition, academic figure understanding, visual assistance, icon-based question answering, synthetic reasoning, and image captioning.
Its training sets contain between 24,000 and 40,000 samples, while each test set contains 3,000 samples.

\begin{table}[H]
\centering
\small
\begin{tabular*}{\columnwidth}{@{\hspace{\tabcolsep}\extracolsep{\fill}}llrr@{\hspace{\tabcolsep}}}
\toprule
\textbf{Dataset} & \textbf{Domain} & \textbf{Train} & \textbf{Test} \\
\midrule
ImageNet-R & Artistic & 24,000 & 3,000 \\
ArxivQA & Academic figures & 40,000 & 3,000 \\
VizWiz & Visual assistance & 40,000 & 3,000 \\
IconQA & Icons & 30,000 & 3,000 \\
CLEVR-Math & Synthetic reasoning & 40,000 & 3,000 \\
Flickr30k & Captioning & 40,000 & 3,000 \\
\bottomrule
\end{tabular*}
\caption{Dataset statistics for the six-task UCIT benchmark.}
\label{tab:ucit_dataset_details}
\end{table}

\section{Derivation of {\moduleB}}
\label{app:routing_derivation}

We derive the coordinate updates used to optimize Eq.~\eqref{eq:kl_argmin}. For a fixed input $\samplex$, we omit its dependence in the intermediate derivation. 
We assume that the relevance scores are strictly positive, as provided by the Gaussian model with positive expert priors.

\tinyparagraph{Expanding the KL objective.}
With the compatibility prior $C_{ij}=\exp(\lambda\mathbb{I}[i=j])$, the target joint distribution is:
\begin{equation}
\Pi_{ij}
=
\frac{
\projrelevance{i}\,
\llmrelevance{j}\,
\exp(\lambda\mathbb{I}[i=j])
}{
Z_{\Pi}
},
\end{equation}
where
$Z_{\Pi}=\sum_{u,v}\projrelevance{u}\llmrelevance{v}
\exp(\lambda\mathbb{I}[u=v])$
is independent of the routing weights. Substituting this expression into the KL divergence gives:
\begin{equation}
\label{eq:app_kl_expansion}
\begin{aligned}
&D_{\mathrm{KL}}\!\left(
q_{\proj}q_{\llm}\,\middle\|\,\Pi
\right)
\\
&=
\sum_{i,j}q_{\proj}^{(i)}q_{\llm}^{(j)}
\log\frac{q_{\proj}^{(i)}q_{\llm}^{(j)}}{\Pi_{ij}}
\\
&=
\sum_{i,j}q_{\proj}^{(i)}q_{\llm}^{(j)}
\Biggl[
\log\frac{q_{\proj}^{(i)}}{\projrelevance{i}}
+
\log\frac{q_{\llm}^{(j)}}{\llmrelevance{j}}
\\
&\hspace{7em}
-\lambda\mathbb{I}[i=j]+\log Z_{\Pi}
\Biggr]
\\
&=
\sum_i q_{\proj}^{(i)}
\log\frac{q_{\proj}^{(i)}}{\projrelevance{i}}
+
\sum_j q_{\llm}^{(j)}
\log\frac{q_{\llm}^{(j)}}{\llmrelevance{j}}
\\
&\quad
-\lambda\sum_i q_{\proj}^{(i)}q_{\llm}^{(i)}
+\log Z_{\Pi}.
\end{aligned}
\end{equation}
The last equality follows from
$\sum_i q_{\proj}^{(i)}=\sum_j q_{\llm}^{(j)}=1$.
Dropping the constant $\log Z_{\Pi}$ recovers the objective $\mathcal{L}$.

\tinyparagraph{Deriving the coordinate updates.}
We minimize $\mathcal{L}$ with respect to one routing distribution while holding the other fixed. This is equivalent to coordinate ascent on the corresponding variational objective, yielding the standard coordinate ascent variational inference procedure~\citep{blei2017variational}.

First, fix $q_{\llm}$ and optimize $q_{\proj}$. Introducing a Lagrange multiplier $\alpha$ for its normalization constraint gives:
\begin{equation}
\begin{aligned}
\mathcal{J}_{\proj}(q_{\proj},\alpha)
={}&
\sum_i q_{\proj}^{(i)}
\log\frac{q_{\proj}^{(i)}}{\projrelevance{i}}
-
\lambda\sum_i q_{\proj}^{(i)}q_{\llm}^{(i)}
\\
&+
\alpha\left(\sum_i q_{\proj}^{(i)}-1\right),
\end{aligned}
\end{equation}
where terms independent of $q_{\proj}$ have been omitted. Setting the derivative with respect to $q_{\proj}^{(i)}$ to zero yields:
\begin{equation}
\frac{\partial\mathcal{J}_{\proj}}
{\partial q_{\proj}^{(i)}}
=
\log\frac{q_{\proj}^{(i)}}{\projrelevance{i}}
+1-\lambda q_{\llm}^{(i)}+\alpha
=
0.
\end{equation}
Rearranging this condition gives:
\begin{equation}
q_{\proj}^{(i)}
=
\projrelevance{i}
\exp\!\left(\lambda q_{\llm}^{(i)}-1-\alpha\right).
\end{equation}
The factor $\exp(-1-\alpha)$ is shared across projector experts and is determined by normalization. Therefore:
\begin{equation}
q_{\proj}^{(i)}
=
\frac{
\projrelevance{i}\exp(\lambda q_{\llm}^{(i)})
}{
\sum_k\projrelevance{k}\exp(\lambda q_{\llm}^{(k)})
}.
\end{equation}
For fixed $q_{\llm}$, the objective is strictly convex in $q_{\proj}$, so this normalized solution is the unique minimizer of the projector coordinate subproblem.

Similarly, fixing $q_{\proj}$ and introducing a multiplier $\beta$ for $\sum_j q_{\llm}^{(j)}=1$ gives the stationarity condition:
\begin{equation}
\log\frac{q_{\llm}^{(j)}}{\llmrelevance{j}}
+1-\lambda q_{\proj}^{(j)}+\beta
=
0.
\end{equation}
Solving and normalizing yields the unique minimizer of the LLM coordinate subproblem:
\begin{equation}
q_{\llm}^{(j)}
=
\frac{
\llmrelevance{j}\exp(\lambda q_{\proj}^{(j)})
}{
\sum_k\llmrelevance{k}\exp(\lambda q_{\proj}^{(k)})
}.
\end{equation}

\tinyparagraph{Sequential routing updates.}
Restoring the input dependence, we initialize both routing distributions with their corresponding relevance scores:
\begin{equation}
\label{eq:app_cavi_init}
\begin{aligned}
q_{\proj}^{(i)}(\samplex;0)
&=\projrelevance{i}(\samplex),\\
q_{\llm}^{(j)}(\samplex;0)
&=\llmrelevance{j}(\samplex).
\end{aligned}
\end{equation}
At iteration $s$, we first update the projector routing distribution and then use its updated values to optimize the LLM routing distribution:
\begin{equation}
\label{eq:app_cavi_update}
\begin{aligned}
q_{\proj}^{(i)}(\samplex;s+1)
&=
\frac{
\projrelevance{i}(\samplex)
e^{\lambda q_{\llm}^{(i)}(\samplex;s)}
}{
\sum_k
\projrelevance{k}(\samplex)
e^{\lambda q_{\llm}^{(k)}(\samplex;s)}
},
\\
q_{\llm}^{(j)}(\samplex;s+1)
&=
\frac{
\llmrelevance{j}(\samplex)
e^{\lambda q_{\proj}^{(j)}(\samplex;s+1)}
}{
\sum_k
\llmrelevance{k}(\samplex)
e^{\lambda q_{\proj}^{(k)}(\samplex;s+1)}
}.
\end{aligned}
\end{equation}
The relevance scores remain fixed throughout these iterations. Each update combines an expert's input relevance with the current routing weight of its paired expert in the other module.

Each coordinate update exactly minimizes its subproblem and therefore cannot increase the KL divergence. Since the KL divergence is bounded below by zero, its values converge if the updates are continued. However, the objective is not generally jointly convex, so these updates do not guarantee a global minimizer of Eq.~\eqref{eq:kl_argmin}. In practice, we perform a fixed number of iterations and use the resulting distributions as the routing weights.

\section{Randomness and Task-Order Robustness}
\label{app:randomness_task_order}

\subsection{Multi-Seed Stability}
\label{app:multi_seed}

To assess run-to-run stability, we rerun the UCIT experiments with five random seeds under otherwise identical settings. We use the original task order: ImageNet-R $\rightarrow$ ArxivQA $\rightarrow$ VizWiz $\rightarrow$ IconQA $\rightarrow$ CLEVR-Math $\rightarrow$ Flickr30k. For every run, we fix the random states of Python, NumPy, PyTorch, CUDA, and the data loader workers, and enable deterministic CUDA operations where available. We compare {\methodshort} with SAME and DISCO, the two strongest baselines in Table~\ref{tab:ucit}. 

\begin{table}[t]
\centering
\small
\begin{tabular}{lccc}
\toprule
\textbf{Seed} & \textbf{SAME} & \textbf{DISCO} & \textbf{\methodshort} \\
\midrule
1 & 72.93 & 68.76 & 75.18 \\
10 & 72.38 & 69.13 & 75.57 \\
100 & 72.56 & 68.81 & 75.75 \\
1000 & 71.86 & 68.50 & 75.70 \\
42 (original) & 73.07 & 69.54 & 75.67 \\
\midrule
\textbf{Average} & 72.56 & 68.95 & 75.58 \\
\textbf{Std.} & 0.48 & 0.40 & 0.23 \\
\bottomrule
\end{tabular}
\caption{Multi-seed results on UCIT. We report final average accuracy (\%) over five random seeds; mean and standard deviation are computed across runs.}
\label{tab:multi_seed}
\end{table}

Table~\ref{tab:multi_seed} shows that {\methodshort} consistently outperforms the strongest baselines across random seeds while exhibiting the smallest run-to-run variation. This confirms that its performance gains are stable and do not depend on a favorable initialization.

\subsection{Task-Order Robustness}
\label{app:task_order}

We further evaluate {\methodshort}'s sensitivity to task order on UCIT. We use seed 42 and keep every other training and evaluation setting identical to the main experiment. We consider three task sequences:
\begin{itemize}
    \item \textbf{Original}: ImageNet-R $\rightarrow$ ArxivQA $\rightarrow$ VizWiz $\rightarrow$ IconQA $\rightarrow$ CLEVR-Math $\rightarrow$ Flickr30k.
    \item \textbf{Reverse}: Flickr30k $\rightarrow$ CLEVR-Math $\rightarrow$ IconQA $\rightarrow$ VizWiz $\rightarrow$ ArxivQA $\rightarrow$ ImageNet-R.
    \item \textbf{Fixed random}: Flickr30k $\rightarrow$ CLEVR-Math $\rightarrow$ ArxivQA $\rightarrow$ ImageNet-R $\rightarrow$ IconQA $\rightarrow$ VizWiz.
\end{itemize}

\begin{table}[t]
\centering
\small
\resizebox{\columnwidth}{!}{%
\begin{tabular}{lccc}
\toprule
\textbf{Metric} & \textbf{Original} & \textbf{Reverse} & \textbf{Fixed random} \\
\midrule
Final Avg. Accuracy (\%) & 75.68 & 74.36 & 75.43 \\
\bottomrule
\end{tabular}
}
\caption{Task-order robustness of {\methodshort} on UCIT. Final average accuracy (\%) is reported under the original, reverse, and fixed random task orders.}
\label{tab:task_order}
\end{table}

As shown in Table~\ref{tab:task_order}, {\methodshort} maintains strong performance under all three task sequences. The limited variation across orders indicates that the method is robust to changes in the chronological arrival of tasks.

\section{Parameter Efficiency}
\label{app:parameter_efficiency}

\begin{figure}[t]
  \centering
  \includegraphics[width=\columnwidth]{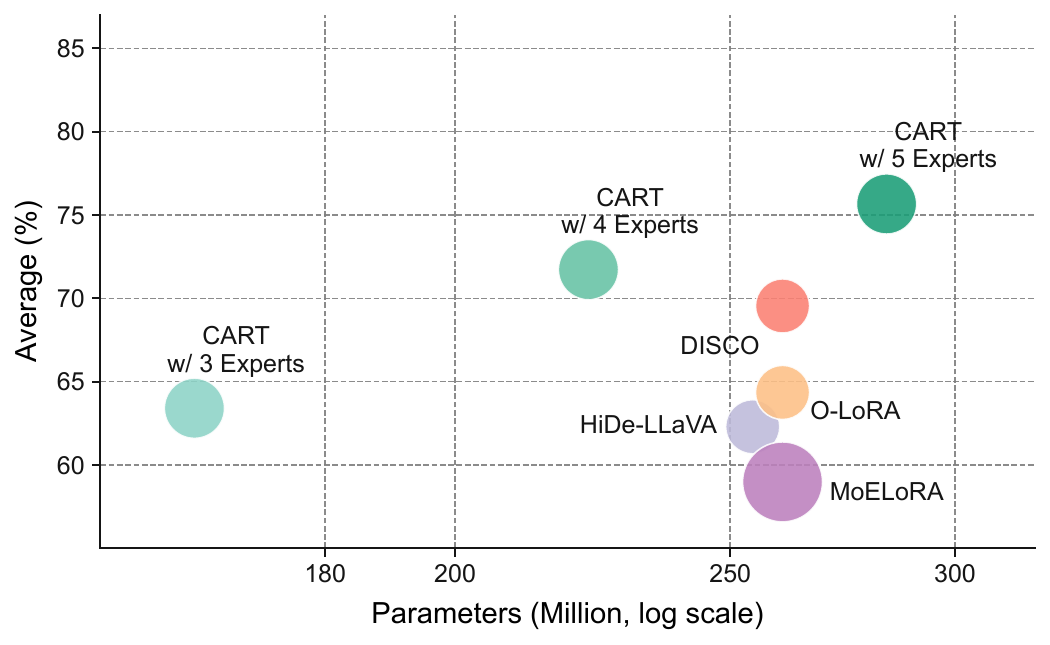}
  \caption{Accuracy--parameter efficiency of MCIT methods on UCIT. {\methodshort} achieves a Pareto-optimal trade-off between Last Accuracy and adaptation parameters.}
  \label{fig:parameter_average}
\end{figure}

We further compare the accuracy--parameter trade-off of MCIT methods on UCIT. In Figure~\ref{fig:parameter_average}, the horizontal axis shows the number of adaptation parameters used at inference, excluding router, mask, and pretrained backbone parameters, while the vertical axis reports \textit{Last Accuracy} (Last), as defined in the Evaluation Metrics paragraph. The radius of each bubble represents the number of trainable parameters introduced for each task. Overall, {\methodshort} achieves a Pareto-optimal trade-off: methods using more adaptation parameters do not attain higher Last Accuracy, whereas methods with higher Last Accuracy require at least as many adaptation parameters.

\section{Pseudocode of {\methodshort}}
\label{sec:pseudocode}
Algorithms~\ref{alg:method_train} and~\ref{alg:method_inference} outline the training and inference procedures of {\methodshort}. For each task, we jointly train a projector expert and an LLM LoRA expert, then freeze the pair and retain its input distribution statistics. At inference time, we use these statistics to estimate input--expert relevance and apply compatibility-aware routing to coordinate the two expert mixtures without requiring task identities.

\begin{algorithm}[t]
\caption{{\methodshort} Training (per task $t$)}
\label{alg:method_train}
\small
\begin{algorithmic}
\Require $\mathcal{D}_t$; frozen $\phi,\psi$; current expert pools
$\boldsymbol{\mathcal{E}}_{\proj},\boldsymbol{\mathcal{E}}_{\llm}$
and their statistics; capacity $K_{\max}$; covariance regularizer
$\epsilon$; LoRA rank $r$
\Ensure Updated paired expert pools $\boldsymbol{\mathcal{E}}_\proj, \boldsymbol{\mathcal{E}}_\llm$ and their distribution statistics
\State Training task-specific experts $\projexpert{t}, \llmexpert{t}$ on $\mathcal{D}_t$ by Eq.~\eqref{eq:mcit_loss}
\State Extract routing representations $\mathbf{h}_\proj(\samplex), \mathbf{h}_\llm(\samplex)$ for $\samplex \in \mathcal{D}_t$ using $\phi, \psi$
\Statex \textbf{Model task distribution}
\For{$m\in\{\proj,\llm\}$}
\State Compute $\boldmu_m^{(t)}$ by Eq.~\eqref{eq:task_mean}, $\mathbf{S}_m^{(t)}$ by Eq.~\eqref{eq:task_std}
\State Update
  $\mathbf{S}_m\gets\mathbf{S}_m+\mathbf{S}_m^{(t)}$
  and
  $\boldSigma_m\gets\mathbf{S}_m/N+\epsilon\mathbf{I}$
\EndFor
\State Append the current experts to the pools $\boldsymbol{\mathcal{E}}_{\proj} \gets \boldsymbol{\mathcal{E}}_{\proj}\cup\{\mathbf{E}_\proj^{(t)}\}, \boldsymbol{\mathcal{E}}_{\llm} \gets \boldsymbol{\mathcal{E}}_{\llm}\cup \{\mathbf{E}_\llm^{(t)}\}$

\Statex \textbf{Expert consolidation}

\If{$|\boldsymbol{\mathcal{E}}_{\proj}|>K_{\max}$}
    \State Compute distribution distances
    $d_{\proj}(i,j)$ and $d_{\llm}(i,j)$
    for all $i<j$ via Eq.~\eqref{eq:distribution_dist}

    \State Select
    $\displaystyle
    (i,j)\gets
    \operatorname*{arg\,min}\nolimits_{i<j}
    \left[
    d_{\proj}(i,j)
    + d_{\llm}(i,j)
    \right]$

    \State Average the parameters of
    $\projexpert{i}$ and $\projexpert{j}$ by Eq.~\eqref{eq:merge_params}

    \State Combine the weight updates of
    $\llmexpert{i}$ and $\llmexpert{j}$ by Eq.~\eqref{eq:merge_params}, retaining LoRA rank $r$ via truncated SVD

    \State Update expert-specific means $\boldmu_\proj^{(\text{new})}, \boldmu_\llm^{(\text{new})}$ by Eq.~\eqref{eq:merge_means}
\EndIf
\end{algorithmic}
\end{algorithm}

\begin{algorithm}
\caption{{\methodshort} Inference}
\label{alg:method_inference}
\small
\begin{algorithmic}
\Require Test input $\samplex=(\mathbf{v},\mathbf{q})$;
trained paired expert pools $\boldsymbol{\mathcal{E}}_\proj, \boldsymbol{\mathcal{E}}_\llm$ and their distribution statistics;
frozen $\phi,\psi$;
compatibility strength $\lambda$;
CAVI iterations $S$
\Ensure Response $\hat{\mathbf{y}}$
\State Extract routing representations $\mathbf{h}_\proj(\samplex), \mathbf{h}_\llm(\samplex)$ using $\phi, \psi$
\State Evaluate expert relevance scores $R_\proj(\samplex), R_\llm(\samplex)$ via Eq.~\eqref{eq:relevance_score}
\Statex \textbf{{\moduleB}}
\State Initialize $q_\proj(\samplex; 0) \gets R_\proj(\samplex)$ and $q_\llm(\samplex; 0) \gets R_\llm(\samplex)$
\For{$s = 0$ to $S-1$}
\State Update $q_\proj, q_\llm$ via Eq.~\eqref{eq:cavi_update}
\EndFor
\State Set $q_\proj(\samplex)\gets q_\proj(\samplex; S)$ and $q_\llm(\samplex)\gets q_\llm(\samplex; S)$

\Statex \textbf{Expert mixture inference}
\State Generate $\hat{\mathbf{y}}$ according to Eq.~\eqref{eq:moe_lora}, using $q_\proj$ and $q_\llm$ as the routing weights 
\end{algorithmic}
\end{algorithm}

\section{Details of Comparison Methods}
\label{app:detail_comparison_method}

\tinyparagraph{EWC}~\citep{kirkpatrick2017overcoming}
assigns greater protection to parameters that are important for earlier tasks. Importance is estimated from Fisher information, and a weighted quadratic penalty discourages these parameters from moving away from their previously learned values during subsequent training.

\tinyparagraph{O-LoRA}~\citep{wang2023orthogonal}
reduces interference by regularizing newly learned low-rank subspaces to remain orthogonal to those associated with earlier tasks. Historical adapters are retained, and their low-rank factors are concatenated for inference, combining task-specific updates within a unified model.

\tinyparagraph{MoELoRA}~\citep{chen2024coin}
replaces a conventional LoRA adapter with several low-rank experts, whose number is configured according to the task sequence. A learned gating network assigns input-dependent weights to these experts, allowing specialized adaptations to contribute differently across tasks.

\tinyparagraph{HiDe-LLaVA}~\citep{guo2025hide}
applies different adaptation strategies at different depths, motivated by layerwise CKA measurements. It expands task-specific LoRA modules in the uppermost transformer layer and retrieves them by matching visual and textual anchors. In lower layers, adapters learned on successive tasks are fused to retain shared knowledge, avoiding the need to optimize a separate routing network.

\tinyparagraph{ProgLoRA}~\citep{yu2025progressive}
grows an adapter pool by allocating a fresh LoRA block to every incoming task and keeping earlier blocks fixed. Its task-aware allocation mechanism combines relevant historical blocks according to inter-task similarity, supporting knowledge transfer during subsequent adaptation. A task recall mechanism further restricts updates to reduce forgetting. The method includes a static variant that assumes access to task identities at test time and a dynamic variant that infers the appropriate allocation when these identities are unavailable.

\tinyparagraph{SAME}~\citep{xie2026same}
addresses forgetting caused by changes in both expert assignments and expert parameters. Its spectral-aware routing decomposes router updates into orthogonal subspaces to balance adaptation with historical consistency. Curvature-aware scaling uses accumulated input covariance to limit changes in expert behavior on earlier tasks. An adaptive activation strategy additionally suspends selected experts during task training, reducing computation and interference.

\tinyparagraph{DISCO}~\citep{guo2025federated}
maintains a dynamic cache of LoRA subspaces for federated continual instruction tuning. Instruction feature prototypes, termed identity tokens, determine how client updates are matched and aggregated within this cache. During inference, subspace selective activation weights each subspace according to the similarity between the input instruction and its identity token, suppressing contributions from unrelated knowledge.

\end{document}